# Learning Parameterized Skills

**Bruno Castro da Silva** BSILVA@CS.UMASS.EDU
Autonomous Learning Laboratory, Computer Science Dept., University of Massachusetts Amherst, 01003 USA.

**George Konidaris** GDK@CSAIL.MIT.EDU
MIT Computer Science and Artificial Intelligence Laboratory, Cambridge MA 02139, USA.

**Andrew G. Barto** BARTO@CS.UMASS.EDU
Autonomous Learning Laboratory, Computer Science Dept., University of Massachusetts Amherst, 01003 USA.

## Abstract

We introduce a method for constructing skills capable of solving tasks drawn from a distribution of parameterized reinforcement learning problems. The method draws example tasks from a distribution of interest and uses the corresponding learned policies to estimate the topology of the lower-dimensional piecewise-smooth manifold on which the skill policies lie. This manifold models how policy parameters change as task parameters vary. The method identifies the number of charts that compose the manifold and then applies non-linear regression in each chart to construct a parameterized skill by predicting policy parameters from task parameters. We evaluate our method on an underactuated simulated robotic arm tasked with learning to accurately throw darts at a parameterized target location.

## 1. Introduction

One approach to dealing with the complexity of applying reinforcement learning to high-dimensional control problems is to specify or discover hierarchically structured policies. The most widely used hierarchical reinforcement learning formalism is the options framework (Sutton et al., 1999), where high-level *options* (also called *skills*) define temporally extended policies that can be used directly in learning and planning but abstract away the details of low-level control. One of the motivating principles underlying hierarchical reinforcement learning is the idea that subproblems recur, so that acquired or designed options can be reused in a variety of tasks and contexts.

However, the options framework as usually formulated defines an option as a *single* policy. An agent may instead wish to define a *parameterized policy* that can be applied across a class of related tasks. For example, consider a soccer playing agent. During a game the agent might wish to kick the ball with varying amounts of force, towards various different locations on the field; for such an agent to be truly competent it should be able to execute such kicks whenever necessary, even with a particular combination of force and target location that it has never had direct experience with. In such cases, learning a single policy for each possible variation of the task is clearly infeasible. The agent might therefore wish to learn good policies for a few specific kicks, and then use this experience to synthesize a single general skill for kicking the ball—parameterized by the amount of force desired and the target location—that it can execute on-demand.

We propose a method for constructing parameterized skills from experience. The agent learns to solve a few instances of the parameterized task and uses these to estimate the topology of the lower-dimensional manifold on which the skill policies lie. This manifold models how policy parameters change as task parameters vary. The method identifies the number of charts that compose the manifold and then applies non-linear regression in each chart to construct a parameterized skill by predicting policy parameters from task parameters. We evaluate the method on an underactuated simulated robotic arm tasked with learning to accurately throw darts at a parameterized target location.

## 2. Setting

In what follows we assume an agent which is presented with a set of tasks drawn from some task distribution. Each task is modeled by a Markov Decision Process (MDP) and the agent must maximize the expected reward over the whole distribution of possible MDPs. We assume that the MDPs have dynamics and reward functions similar enough so that they can be considered variations of a same task. Formally, the goal of such an agent is to maximize:

$$\int P(\tau) J\big(\pi_\theta, \tau\big) d\tau, \tag{1}$$

where $\pi_\theta$ is a policy parameterized by a vector $\theta \in \mathbb{R}^N$, $\tau$ is a task parameter vector drawn from a $|T|$-dimensional continuous space $T$, $J(\pi, \tau) = E\big\{ \sum_{t=0}^{K} r_t | \pi, \tau \big\}$ is the expected return obtained when executing policy $\pi$ while in task $\tau$ and $P(\tau)$ is a probability density function describing the probability of task $\tau$ occurring. Furthermore, we define a *parameterized skill* as a function

$$\Theta : T \to \mathbb{R}^N,$$

mapping task parameters to policy parameters. When using a parameterized skill to solve a distribution of tasks, the specific policy parameters to be used depend on the task currently being solved and are specified by $\Theta$. Under this definition, our goal is to construct a parameterized skill $\Theta$ which maximizes:

$$\int P(\tau) J\big(\pi_{\Theta(\tau)}, \tau\big) d\tau. \tag{2}$$

### 2.1. Assumptions

We assume the agent must solve tasks drawn from a distribution $P(\tau)$. Suppose we are given a set $K$ of pairs $\{\tau, \theta_\tau\}$, where $\tau$ is a $|T|$-dimensional vector of task parameters sampled from $P(\tau)$ and $\theta_\tau$ is the corresponding policy parameter vector that maximizes return for task $\tau$. We would like to use $K$ to construct a parameterized skill which (at least approximately) maximizes the quantity in Equation 2.

We start by highlighting the fact that the probability density function $P$ induces a (possibly infinite) set of skill policies for solving tasks in the support of $P$, each one corresponding to a vector $\theta_\tau \in \mathbb{R}^N$. These policies lie in an $N$-dimensional space containing sample policies that can be used to solve tasks drawn from $P$. Since the tasks in the support of $P$ are assumed to be related, it is reasonable to further assume that there exists some structure in this space; specifically, that the policies for solving tasks drawn from the distribution lie on a lower-dimensional surface embedded in $\mathbb{R}^N$ and that their parameters vary smoothly as we vary the task parameters.

This assumption is reasonable in a variety of situations, especially in the common case where the policy is differentiable with respect to its parameters. In this case, the natural gradient of the performance $J\big(\pi_\theta, \tau\big)$ is well-defined and indicates the direction (in policy space) that locally maximizes $J$ but which does not change the distribution of paths induced by the policy by much. Consider, for example, problems in which performance is directly correlated to how close the agent gets to a goal state; in this case one can interpret a small perturbation to the policy as defining a *new* policy which solves a similar task but with a slightly different goal. Since under these conditions small policy changes induce a smoothly-varying set of goals, one can imagine that the goals themselves parameterize the space of policies: that is, that by varying the goal or task one moves over the lower-dimensional surface of corresponding policies.

Note that it is possible to find points in policy space in which the corresponding policy cannot be further locally modified in order to obtain a solution to a new, related goal. This implies that the set of skill policies of interest might be in fact distributed over several charts of a piecewise-smooth manifold. Our method can automatically detect when this is the case and construct separate models for each manifold, essentially discovering how many different skills exist and creating a unified model by which they are integrated.

## 3. Overview

Our method proceeds by collecting example task instances and their solution policies and using them to train a family of independent non-linear regression models mapping task parameters to policy parameters. However, because policies for different subsets of $T$ might lie in different, disjoint manifolds, it is necessary to first estimate how many such lower-dimensional surfaces exist before separately training a set of regression models for each one.

More formally, our method consists of four steps: *1)* draw $|K|$ sample tasks from $P$ and construct $K$, the set of task instances $\tau$ and their corresponding learned policy parameters $\theta_\tau$; *2)* use $K$ to estimate the geometry and topology of the policy space, specifically the number $D$ of lower-dimensional surfaces embedded in $\mathbb{R}^N$ on which skill policies lie; *3)* train a classifier $\chi$ mapping elements of $T$ to $[1, \ldots, D]$; that is, to one of the $D$ lower-dimensional manifolds; *4)* train a set of $(N \times D)$ independent non-linear regression models $\Phi_{i,j}$, $i \in [1, \ldots, D]$, $j \in [1, \ldots N]$, each one mapping elements of $T$ to individual skill policy parameters $\theta_i$, $i \in [1, \ldots N]$. Each subset $[\Phi_{i,1}, \ldots, \Phi_{i,N}]$ of regression models is trained over all tasks $\tau$ in $K$ where $\chi(\tau) = i$.[1] We therefore define a parameterized skill as a vector function:

$$\Theta(\tau) \equiv [\Phi_{\chi(\tau),1}, \ldots, \Phi_{\chi(\tau),N}]^T. \quad (3)$$

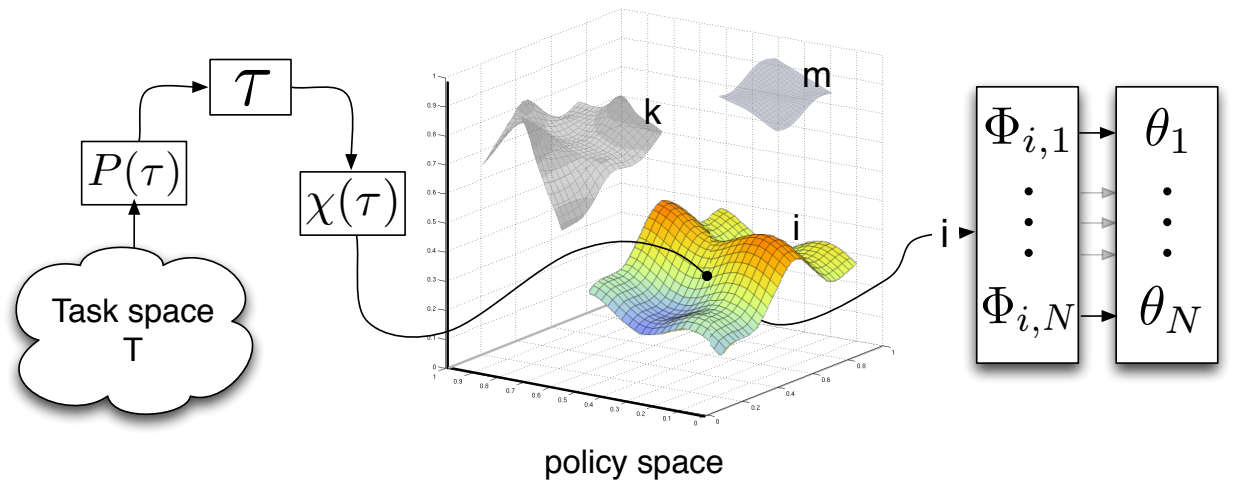

*Figure 1.* Steps involved in executing a parameterized skill: a task is drawn from the distribution $P$; the classifier $\chi$ identifies the manifold to which the policy for that task belongs; the corresponding regression models for that manifold map task parameters to policy parameters.

Figure 1 depicts the above-mentioned steps. Note that we have described our method without specifying a particular choice of policy representation, learning algorithm, classifier, or non-linear regression model, since these design decisions are best made in light of the characteristics of the application at hand. In the following sections we present a control problem whose goal is to accurately throw darts at a variety of targets and describe one possible instantiation of our approach.

# 4. The Dart Throwing Domain

In the dart throwing domain, a simulated planar underactuated robotic arm is tasked with learning a parameterized policy to accurately throw darts at targets around it (Figure 4). The base of the arm is affixed to a wall in the center of a 3-meter high and 4-meter wide room. The arm is composed of three connected links and a single motor which applies torque only to the second joint, making this a difficult non-linear and underactuated control problem. At the end of its third link, the arm possesses an actuator capable of holding and releasing a dart. The state of the system is a 7-dimensional vector composed by 6 continuous features corresponding to the angle and angular velocities of each link and by a seventh binary feature specifying whether or not the dart is still in being held. The goal of the system is to control the arm so that it executes a throwing movement and accurately hits a target of interest. In this domain the space $T$ of tasks consists of a 2-dimensional Euclidean space containing all $(x, y)$ coordinates at which a target can be placed—a target can be affixed anywhere on the walls or ceiling surrounding the agent.

[1] This last step assumes that the policy features are approximately independent conditioned on the task; if this is known not to be the case, it is possible to alternatively train a set of $D$ multivariate non-linear regression models $\Phi_i$, $i \in [1, \ldots, D]$, each one mapping elements of $T$ to complete policies parameterizations $\theta \in \mathbb{R}^N$, and use them to construct $\Theta$. Again, the $i$-th such model should be trained only over tasks $\tau$ in $K$ such that $\chi(\tau) = i$.

# 5. Learning Parameterized Skills for Dart Throwing

To implement the method outlined in Section 3 we need to specify methods to *1)* represent a policy; *2)* learn a policy from experience; *3)* analyze the topology of the policy space and estimate $D$, the number of lower-dimensional surfaces on which skill policies lie; *4)* construct the non-linear classifier $\chi$; and *5)* construct the non-linear regression models $\Phi$. In this section we describe the specific algorithms and techniques chosen in order to tackle the dart-throwing domain. We discuss our results in Section 6.

Our choices of methods are directly guided by the characteristics of the domain. Because the following experiments involve a multi-joint simulated robotic arm, we chose a policy representation that is particularly well-suited to robotics: Dynamic Movement Primitives (Schaal et al., 2004), or DMPs. DMPs are a framework for modular motor control based on a set of linearly-parameterized autonomous non-linear differential equations. The time evolution of these equations defines a smooth kinematic control policy which can be used to drive the controlled system. The specific trajectory in joint space that needs to be followed is obtained by integrating the following set of differential equations:

$$\kappa \dot{v} = K(g - x) - Qv + (g - x_0)f$$

$$\kappa \dot{x} = v,$$

where $x$ and $v$ are the position and velocity of the system, respectively; $x_0$ and $g$ denote the start and goal positions; $\kappa$ is a temporal scaling factor; and $K$ and $Q$ act like a spring constant and a damping term, respectively. Finally, $f$ is a non-linear function which can be learned in order to allow the system to generate arbitrarily complex movements and is defined as

$$f(s) = \frac{\sum_i w_i \psi_i(s)}{\sum_i \psi_i(s)},$$

where $\psi_i(s) = \exp(-h_i(s - c_i)^2)$ are Gaussian basis functions with adjustable weights $w_i$ and which depend on a phase variable $s$. The phase variable is constructed so that it monotonically decreases from 1 to 0 during the execution of the movement and is typically computed by integrating $\kappa \dot{s} = -\alpha s$, where $\alpha$ is a pre-determined constant. In our experiments we used a PID controller to track the trajectories induced by the above-mentioned system of equations.

This results in a 37-dimensional policy vector $\theta = [\lambda, g, w_1, \ldots, w_{35}]^T$, where $\lambda$ specifies the value of the phase variable $s$ at which the arm should let go of the dart; $g$ is the goal parameter of the DMP; and $w_1, \ldots, w_{35}$ are the weights of each Gaussian basis function in the movement primitive.

We combine DMPs with PoWER (Kober & Peters, 2008), a policy search technique that collects sample path executions and updates the policy's parameters towards ones that induce a new success-weighted path distribution. PoWER works by executing rollouts $\rho$ constructed based on slightly perturbed versions of the current policy parameters; perturbations to the policy parameters consist of a structured, state-dependent exploration ${\varepsilon_\mathbf{t}}^T \phi(\mathbf{s}, t)$, where $\varepsilon_\mathbf{t} \sim \mathcal{N}(0, \hat{\mathbf{\Sigma}})$ and $\hat{\mathbf{\Sigma}}$ is a meta-parameter of the exploration; $\phi(\mathbf{s}, t)$ is the vector of policy feature activations at time $t$. By adding this type of perturbation to $\theta$ we induce a stochastic policy whose actions are $a = (\theta + \varepsilon_\mathbf{t})^T \phi(\mathbf{s}, t)) \sim \mathcal{N}(0, \phi(\mathbf{s}, t)^T \hat{\mathbf{\Sigma}} \phi(\mathbf{s}, t))$. After performing rollouts using such a stochastic policy, the policy parameters are updated as follows:

$$\theta_{k+1} = \theta_k + \left\langle \sum_{t=1}^{T} \mathbf{W}(\mathbf{s}, t) Q^\pi(\mathbf{s}, \mathbf{a}, t)) \right\rangle_{\omega(\rho)}^{-1} \times \left\langle \sum_{t=1}^{T} \mathbf{W}(\mathbf{s}, t) \varepsilon_t Q^\pi(\mathbf{s}, \mathbf{a}, t)) \right\rangle_{\omega(\rho)}$$

where $\hat{Q}^\pi(\mathbf{s}, \mathbf{a}, t) = \sum_{\tilde{t}=t}^{T} r(\mathbf{s}_{\tilde{t}}, \mathbf{a}_{\tilde{t}}, \mathbf{s}_{\tilde{t}+1}, \tilde{t})$ is an unbiased estimate of the return, $\mathbf{W}(\mathbf{s}, t) = \phi(\mathbf{s}, t)\phi(\mathbf{s}, t)^T \left(\phi(\mathbf{s}, t)^T \hat{\mathbf{\Sigma}} \phi(\mathbf{s}, t)\right)^{-1}$ and $\langle \cdot \rangle_{\omega}(\rho)$ denotes an importance sampler which can be chosen depending on the domain. A useful heuristic when defining $\omega$ is to discard sample rollouts with very small importance weights; importance weights, in our experiments, are proportional to the relative performance of the rollout in comparison to others.

To analyze the geometry and topology of the policy space and estimate the number $D$ of lower-dimensional surfaces on which skill policies lie we used the ISOMAP algorithm (Tenenbaum et al., 2000). ISOMAP is a technique for learning the underlying global geometry of high-dimensional spaces and the number of non-linear degrees of freedom that underlie it. This information provides us with an estimate of $D$, the number of disjoint lower-dimensional manifolds where policies are located; ISOMAP also specifies to which of these disconnected manifolds a given input policy belongs. This information is used to train the classifier $\chi$, which learns a mapping from task parameters to numerical identifiers specifying one of the lower-dimensional surfaces embedded in policy space. For this domain we have implemented $\chi$ by means of a simple linear classifier. In general, however, more powerful classifiers could be used.

Finally, we must choose a non-linear regression algorithm for constructing $\Phi_{i,j}$. We use standard Support Vector Machines (SVM) (Vapnik, 1995) due to their good generalization capabilities and relatively low dependence on parameter tuning. In the experiments presented in Section 6 we use SVMs with Gaussian kernels and a inverse variance width of 5.0. As previously mentioned, if important correlations between policy and task parameters are known to exist, multivariate regression models might be preferable; one possibility in such cases are Structure Support Vector Machines (Tsochantaridis et al., 2005).

## 6. Experiments

Before discussing the performance of parameterized skill learning in this domain, we present some empirically measured properties of its policy space. Specifically, we describe topological characteristics of the induced space of policies generated as we vary the task. We sampled 60 tasks (target positions) uniformly at random and placed target boards at the corresponding positions. Policies for solving each one of these tasks were computed using PoWER; a policy update was performed every 20 rollouts and the search ran until a *minimum performance threshold* was reached. In our simulations, this criteria corresponded to the moment when the robotic arm first executed a policy that landed the dart within 5 centimeters of the intended target. In order to speed up the sampling process we initialize policies for subsequent targets with ones computed for previously sampled tasks.

We first analyze the structure of the policy manifold by estimating how each dimension of a policy varies as we smoothly vary the task. Figure 2a presents this information for a representative subset of policy parameters. On each subgraph of Figure 2a the $x$ axis corresponds to a 1-dimensional representation of the task obtained by computing the angle at which the target is located with respect to the arm; this is done for ease of visualization, since using $x, y$ coordinates would require a 3-D figure. The $y$ axis corresponds to the value of a selected policy parameter. The first important observation to be made is that as we vary the task, not only do the policy parameters vary smoothly, but they tend to remain confined to one of two disjoint but smoothly varying lower-dimensional surfaces. A discontinuity exists, indicating that after a certain point in task space a qualitatively different type of policy parameterization is required. Another interesting observation is that this discontinuity occurs approximately at the task parameter values corresponding to hitting targets directly above the robotic arm; this implies that skills for hitting targets to the left of the arm lie on a different manifold than policies for hitting targets to its right. This information is relevant for two reasons: *1)* it confirms both that the manifold assumption is reasonable and that smooth task variations induce smooth, albeit non-linear, policy changes; and *2)* it shows that the policies for solving a distribution of tasks are generally confined to one of several lower-dimensional surfaces, and that the way in which they are distributed among these surfaces is correlated to the qualitatively different strategies that they implement.

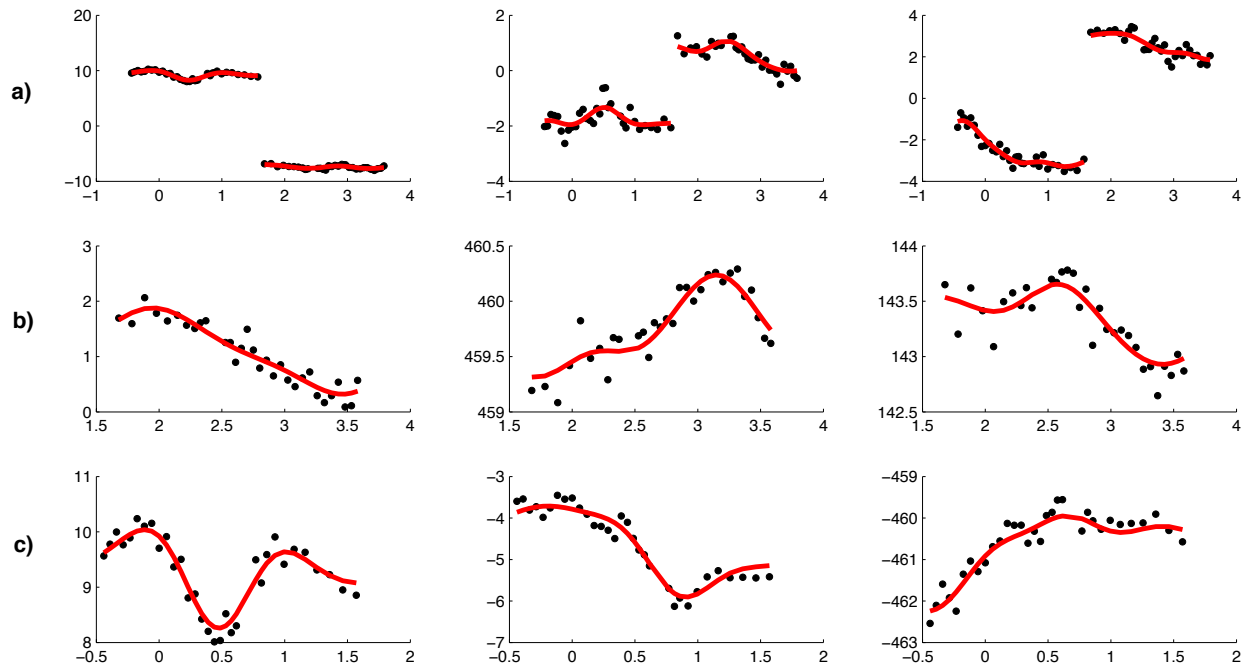

*Figure 2.* Analysis of the variation of a subset of policy parameters as a function of smooth changes in the task.

Figures 2b and 2c show, similarly, how a selected subset of policy parameters changes as we vary the task, but now with the two resulting manifolds analyzed separately. Figure 2b shows the variations in policy parameters induced by smoothly modifying tasks for hitting targets anywhere in the interval of 1.57 to 3.5 radians—that is, targets placed roughly at angles between 90° (directly above the agent) and 200° (lowest part of the right wall). Figure 2c shows that same information but for targets located on one of the other two quadrants—that is, targets to the left of the arm.

We superimposed in Figures 2a-c a red curve representing the non-linear fit constructed by $\Phi$ while modeling the relation between task and policy parameters in each manifold. Note also how a clear linear separation exists between which task policies lie on which manifold: this separation indicates that two qualitatively distinct types of movement are required for solving different subsets of the tasks. Because we empirically observe that a linear separation exists, we implement $\chi$ using a simple linear classifier mapping tasks parameters to the numerical identifier of the manifold to which the task belongs.

We can also analyze the characteristics of the lower-dimensional, quasi-isometric embedding of policies produced by ISOMAP. Figure 3 shows the 2-dimensional embedding of a set of policies sampled from one of the manifolds. Embeddings for the other manifold have similar properties. Analysis of the residual variance of ISOMAP allows us to conclude that the intrinsic dimensionality of the skill manifold is 2; this is expected since we are essentially parameterizing a high-dimensional policy space by task parameters, which are drawn from the 2-dimensional space $T$. This implies that even though skill policies themselves are part of a 37-dimensional space, because there are just two degrees-of-freedom with which we can vary tasks, the policies themselves remain confined to a 2-dimensional manifold. In Figure 3 we use lighter-colored points to identify embeddings of policies for hitting targets at higher locations. From this observation it is possible to note how policies for similar tasks tend to remain geometrically close in the space of solutions.

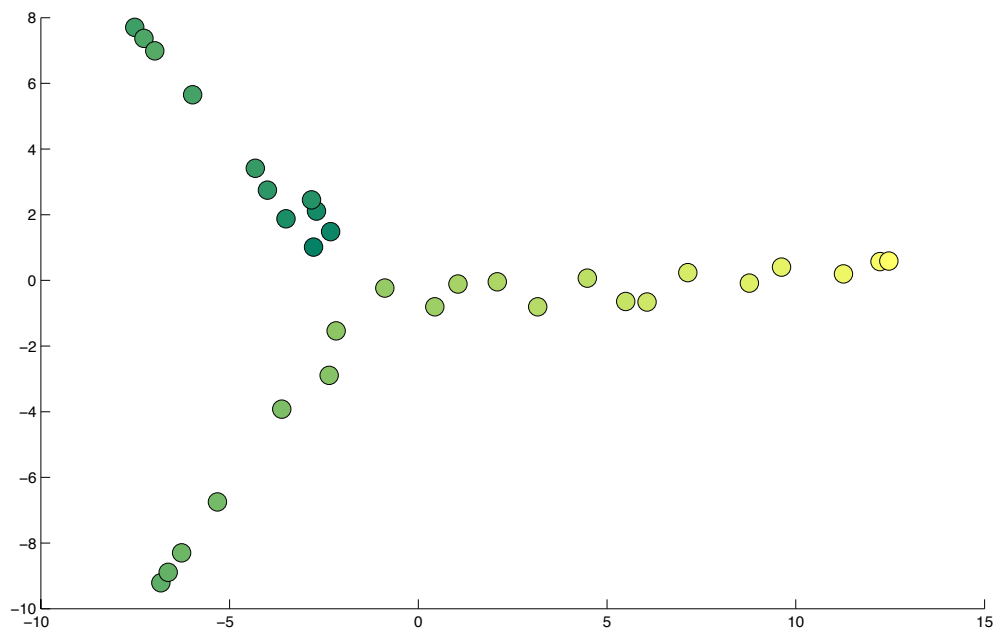

*Figure 3.* 2-dimensional embedding of policies parameters.

Figure 4 shows some types of movements the arm is capable of executing when throwing the dart at specific targets. Figure 4a and Figure 4b present trajectories corresponding to policies aiming at targets high on the ceiling and low on the right wall, respectively; these were presented as training examples to the parameterized skill. Note that the link trajectories required to accurately hit a target are complex because we are using just a single actuated joint to control an arm with three joints.

Figure 4c shows a policy that was predicted by the parameterized skill for a new, unknown task corresponding to a target in the middle of the right wall. A total of five sample trajectories were presented to the parameterized skill and the corresponding predicted policy was further improved by two policy updates, after which the arm was capable of hitting the intended target perfectly.

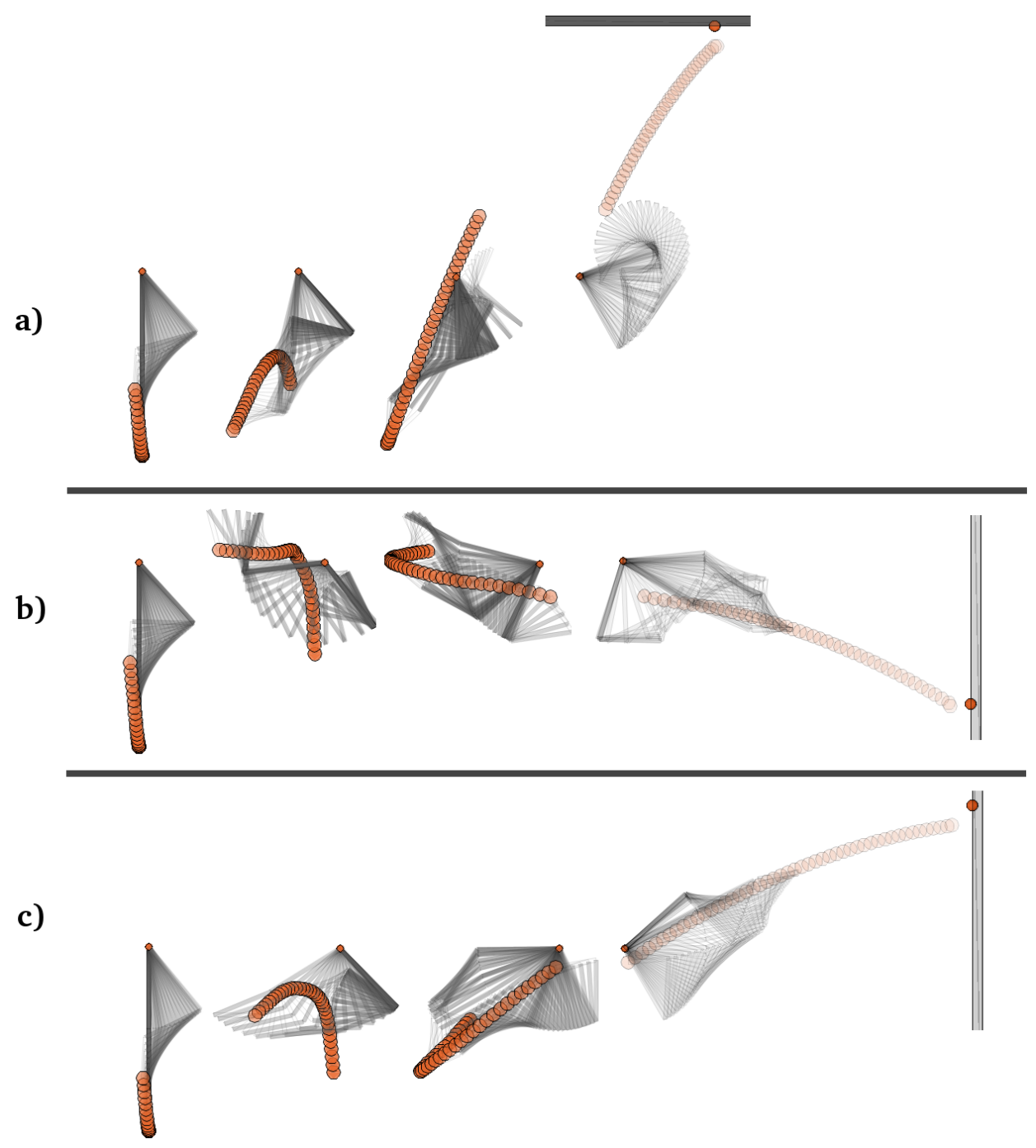

*Figure 4.* Learned arm movements (a,b) presented as training examples to the parameterized skill; (c) predicted movement for a novel target.

Figure 5 shows the predicted policy parameter error, averaged over the parameters of 15 unknown tasks sampled uniformly at random, as a function of the number of examples used to learn the parameterized skill. This is a measure of the relative error between the policy parameters predicted by $\Theta$ and parameters of a known good solution for the same task. The lower the error, the closer the predicted policy is (in norm) to the correct solution. After 6 samples are presented to the parameterized option it is capable of predicting policies whose parameters are within 6% of the correct ones; with approximately 15 samples, this error stabilizes around 3%. Note that this type of accuracy is only possible because even though the spaces analyzed are high-dimensional, they are also highly structured; specifically, solutions to similar tasks lie on a lower-dimensional manifold whose regular topology can be exploited when generalizing known solutions to new problems.

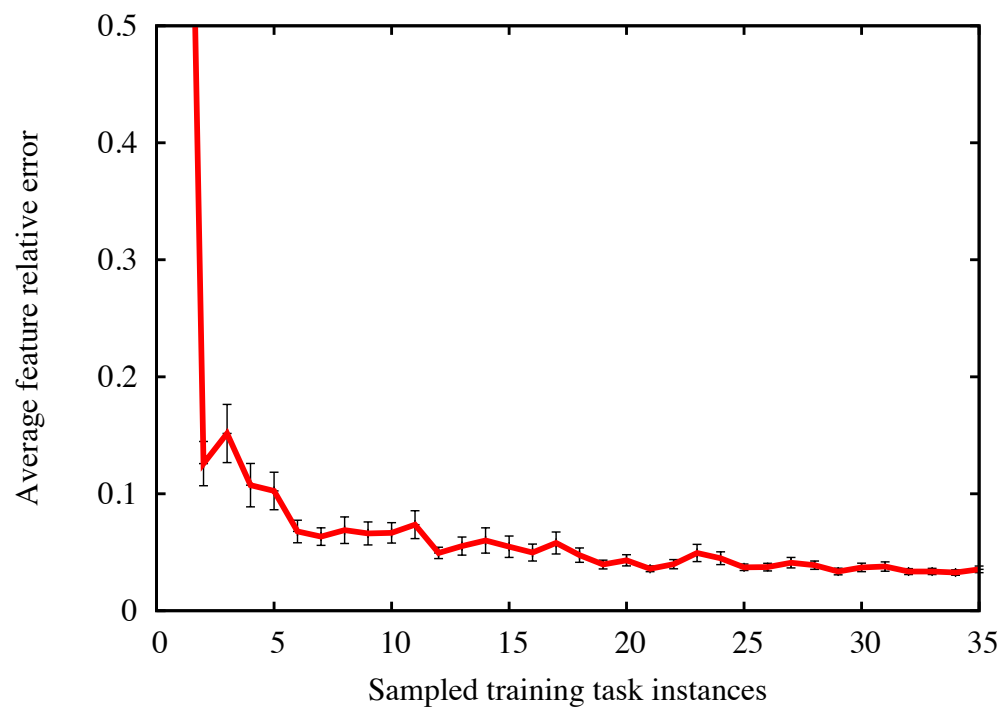

*Figure 5.* Average predicted policy parameter error as a function of the number of sampled training tasks.

Since some policy representations might be particularly sensitive to noise, we additionally measured the actual effectiveness of the predicted policy when directly applied to novel tasks. Specifically, we measure the distance between the position where the dart hits and the intended target; this measurement is obtained by executing the predicted policy directly and *before* any further learning takes places. Figure 6 shows that after 10 samples are presented to the parameterized skill, the average distance is 70cm. This is a reasonable error if we consider that targets can be placed anywhere on a surface that extends for a total of 10 meters. If the parameterized skill is presented with a total of 24 samples the average error decreases to 30cm, which roughly corresponds to the dart being thrown from 2 meters away and landing one dartboard away from the intended center.

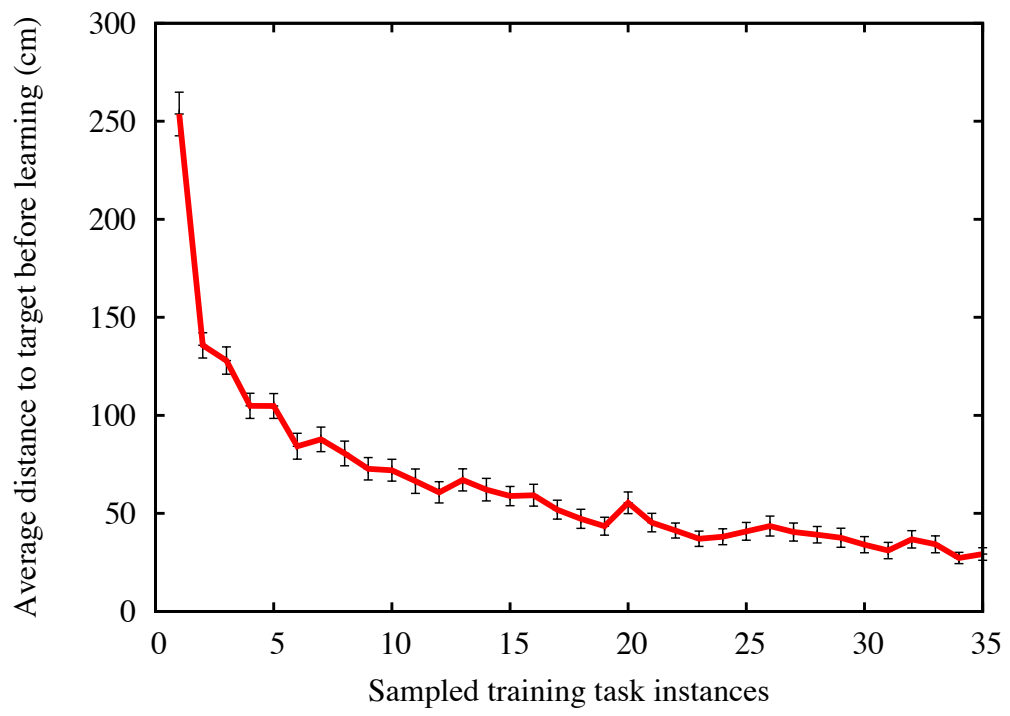

*Figure 6.* Average distance to target (before learning) as a function of the number of sampled training tasks.

Although these initial solutions are good, especially considering that no learning with the target task pa-rameters took place, they are not perfect. We might therefore want to further improve them. Figure 7 shows how many additional policy updates are required to improve the policy predicted by the parameterized skill up to a point where it reaches a performance threshold. The dashed line in Figure 7 shows that on average 22 policy updates are required for finding a good policy when the agent has to learn from scratch. On the other hand, by using a parameterized skill trained with 9 examples it is already possible to decrease this number to just 4 policy updates. With 20 examples or more it takes the agent an average of 2 additional policy updates to meet the performance threshold.

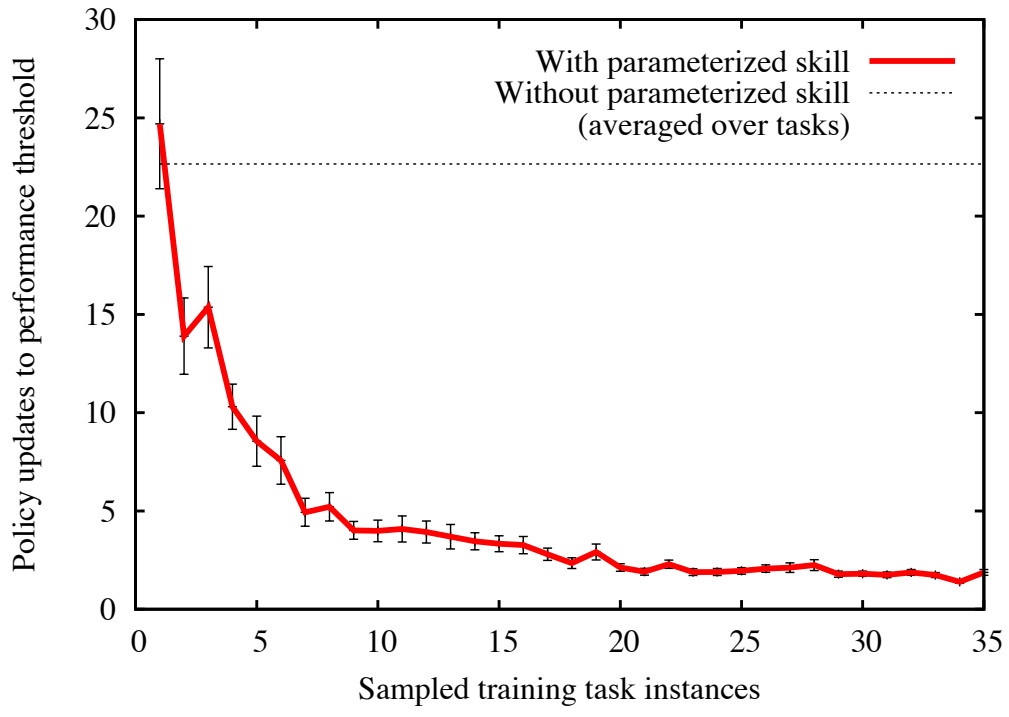

*Figure 7.* Average number of policy updates required to improve the solution predicted by the parameterized skill as a function of the number of sampled training tasks.

## 7. Related Work

The simplest solution for learning a distribution of tasks in RL is to include $\tau$, the task parameter vector, as part of state descriptor and treat the entire class of tasks as a single MDP. This approach has several shortcomings: *1)* learning and generalizing over tasks is slow since the state features corresponding to task parameters remain constant throughout episodes; *2)* the number of basis functions required to approximate the value function or policy needs to be increased to capture all the non-trivial correlations caused by the added task features; *3)* sample task policies cannot be collected in parallel and combined to accelerate the construction of the skill; and *4)* if the distribution of tasks is non-stationary, there is no simple way of adapting a single estimated policy in order to deal with a new pattern of tasks.

Alternative approaches have been proposed under the general heading of skill transfer. Konidaris and Barto (2007) learn reusable options by representing them in an agent-centered state space but do not address the problem of how to construct skills for solving a family of related tasks. Soni and Singh (2006) create options whose termination criteria can be adapted on-the-fly to deal with changing aspects of a task. They do not, however, predict a complete parameterization of the policy for new tasks.

Other methods have been proposed to transfer a model or value function between given pairs of tasks, but not necessarily to reuse learned tasks and construct a parameterized solution. It is often assumed that a mapping between features and actions of a source and target tasks exists and is known *a priori*, as in Taylor and Stone (2007). Liu and Stone (2006) propose a method for transferring a value function between pairs of tasks but require prior knowledge of the task dynamics in the form of a Dynamic Bayes Network. Hausknecht and Stone (2011) present a method for learning a parameterized skill for kicking a ball with varying amounts of force. They exhaustively test variations of one of the policy parameters known *a priori* to be relevant for the skill and then measure the resulting effect on the distance traveled by the ball. By assuming a quadratic relation between these variables, they are able to construct a regression model and invert it, thereby obtaining a closed-form expression for the policy parameter value required for a desired kick. This is an interesting example of the type of parameterized skill that we would like to construct, albeit a very domain-dependent one.

The work most closely related to ours is by Kober, Wilhelm, Oztop, and Peters (2012), who learn a mapping from task description to metaparameters of a DMP, estimating the mean value of each metaparameter given a task and the uncertainty with which it solves that task. Their method uses a parameterized skill framework similar to ours but requires the use of DMPs for policies and assumes that their metaparameters are sufficient to represent the class of tasks of interest. Bitzer, Havoutis, and Vijayakumar (2008) synthesize novel movements by modulating DMPs learned in a latent space and projecting them back onto the original pose space. Neither approach supports arbitrary task parameterizations or discontinuities in the skill manifold, which are important, for instance, for classes of movements whose description requires more than one DMP.

Finally, Braun et al. (2010) discuss how Bayesian modeling can be used to explain experimental data from cognitive and motor neuroscience that supports the idea of structure learning in humans, a concept very similar to the one of parameterized skills. The authors do not, however, propose a concrete method for iden-tifying and constructing such skills.

## 8. Conclusions and Future Work

We have presented a general framework for constructing parameterized skills. The idea underlying our method is to sample a small number of task instances and generalize them to new problems by combining classifiers and non-linear regression models. This approach is effective in practice because it exploits the intrinsic structure of the policy space and because skill policies for similar tasks typically lie on a lower-dimensional manifold. Our framework allows for the construction of effective parameterized skills and is able to identify the number of qualitatively different strategies required for a given distribution of tasks.

This work can be extended in several important directions. First, the question of how to actively select training tasks in order to improve the overall readiness of a parameterized skill, given a distribution of tasks expected in the future, needs to be addressed. Another important open problem is how to properly deal with a non-stationary distribution of tasks. If a new task distribution is known exactly it might be possible to use it to resample instances from $K$ and thus reconstruct the parameterized skill. However, more general strategies are needed if the task distribution changes in a way that is not known to the agent.

Another important question is how to analyze the topology and geometry of the policy space more efficiently. Methods for discovering the underlying global geometry of high-dimensional spaces typically require dense sampling of the manifold, which could require solving an unreasonable number of training tasks. Note, however, that most local policy search methods like the Natural Actor Critic and PoWER move smoothly over the manifold of policies while searching for locally optimal solutions. Therefore, at each policy update during learning they provide us with a new sample which can be used for further train the parameterized skill; each task instance therefore results in a *trajectory* through policy parameter space. Integrating this type of sampling into the construction of the skill corresponds to a type of off-policy learning method, since samples collected while estimating one policy could be used to generalize it to different tasks.

## Acknowledgments

This research was partially funded by the European Union under the FP7-ICT program (IM-CLeVeR project, grant agreement no. ICT-IP-231722).